%% file: root.tex
\documentclass[letterpaper, 10 pt, conference]{./support/ieeeconf}
\usepackage{times}
\usepackage[pdftex]{graphicx}
\usepackage[font={footnotesize}]{caption}
\usepackage{subcaption}

\usepackage{multicol}
\usepackage{amsmath,amssymb,amsopn,amstext,amsfonts}
\usepackage{url}
\usepackage{balance}
\usepackage[usenames, dvipsnames]{color}
\usepackage[linkcolor=black,citecolor=black,urlcolor=black,colorlinks=true,bookmarks=true]{hyperref}
\usepackage[capitalise]{cleveref} 
\usepackage[table]{xcolor}
\usepackage{multirow}
\usepackage{mathrsfs}
\usepackage{isomath}
\usepackage{gensymb}
\usepackage{verbatim}
\usepackage{makecell}

\usepackage{booktabs} 
\usepackage{tikz} 
\usetikzlibrary{shapes,arrows} 

\usepackage[linesnumbered,ruled,lined,vlined]{algorithm2e}

\SetCommentSty{mycommfont}
\SetAlFnt{\small}

\IEEEoverridecommandlockouts
\usepackage{./support/macro}

\makeatletter
\let\NAT@parse\undefined
\makeatother
\usepackage[numbers,sort&compress]{natbib}

\DeclareGraphicsExtensions{.pdf,.png,.jpg}
\graphicspath{{figures/}}

\title{\LARGE \bf
An imminent collision monitoring system with safe stopping interventions for autonomous aerial flights
}

\author{Jasmine Cheng, Xuning Yang and Nathan Michael
\thanks{The Authors are with the Robotics Institute at Carnegie Mellon University, Pittsburgh PA, 15213, USA {\tt \{xuning, nmichael\}@cmu.edu}}%
}

\begin{document}

\maketitle
\thispagestyle{empty}
\pagestyle{empty}

\begin{abstract}
\input{abstract}
\end{abstract}

\input{introduction}
\input{methodology}

\input{fig_warehouse_run.tex}

\input{experiments}
\input{conclusion}
\input{fig_forest_hw_run.tex}

{\scriptsize
\nocite{*}
\bibliographystyle{unsrtnat} 
\bibliography{references}
}

\end{document}

%% file: abstract.tex
Collision avoidance requires tradeoffs in planning time horizons. Depending on the planner, safety cannot always be guaranteed in uncertain environments given map updates. To mitigate situations where the planner leads the vehicle into a state of collision or the vehicle reaches a point where no trajectories are feasible, we propose a continuous collision checking algorithm. The imminent collision checking system continuously monitors vehicle safety, and plans a safe trajectory that leads the vehicle to a stop within the observed map. We test our proposed pipeline alongside a teleoperated navigation in real-life experiments, and in simulated random-forest and warehouse environments where we show that with our method, we are able to mitigate collisions with a success rate of at least 90\%.

%% file: introduction.tex
\section{Introduction}

In tasks such as navigation and exploration in dense environments, static and dynamic collision avoidance require trade-offs in planning time horizons.
Planning at a high frequency results in increased demand for computation at unrealistic rates, while planning at low frequencies may result in inaccurate plans if map updates between planning iterations.
This work seeks to improve aerial vehicle resilience and robustness in a variety of environments by introducing a safety monitoring system that can supplement any autonomous or semi-autonomous aerial system.
Prior work has explored collision avoidance and replanning with respect to goal-seeking, where the vehicle is autonomously controlled and the goal is known a priori \cite{FERREIRA2008}. 
The proposed method mitigates possible collisions at map update rate in a lightweight way using the vehicle state, preventing the vehicle from reaching an unrecoverable state in situations where the planner fails to mitigate the collision in time.

The proposed system prevents collisions by continuously monitoring collision safety and generating a stopping trajectory if the vehicle is at risk of collision.
This system is first included in \cite{Yang2021icra}. In this work, we demonstrate the imminent collision monitoring system in a variety of scenarios and show its applicability and robustness with respect to a wide variety of vehicle and environment conditions. We show that, with the inclusion of the proposed system across a variety of autonomous/semi-autonomous flights, we are able to detect unsafe states and safely avoid collisions.

%% file: methodology.tex
\section{Method}

The proposed method proceeds as follows: the system continuously monitors collision possibility by evaluating the closest obstacles to the vehicle's current location at a fixed rate. Then, we generate a dynamically feasible and safe trajectory and bring the vehicle to a stop by first selecting an escape point along the direction of motion, then generating a trajectory to the escape point. The method is detailed as follows.
\subsection{Imminent collision checking}

\input{collision_monitoring.tex}

\input{fig_open.tex}
\begin{figure*}[t!]
  \centering
  \includegraphics[width=0.95\linewidth]{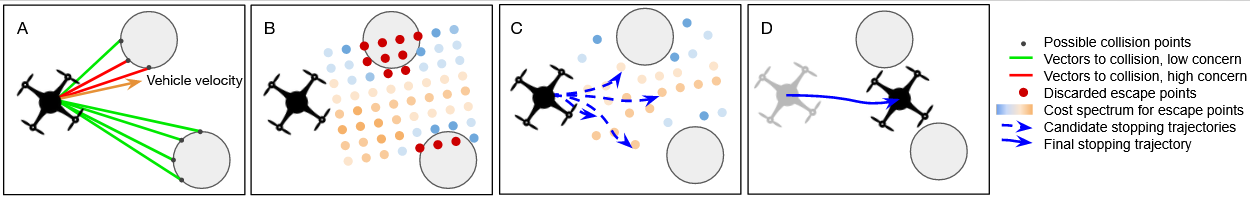}
	\caption{Illustration of imminent collision monitoring and safe stop. (A) Initial assessment: possible collision points are sampled from nearby obstacles and vectors to collision are assessed. If high concern vectors exist, stopping trajectory generation begins. (B) Initial grid of escape points with costs computed and colliding points discarded (C) A stratified sampling method is used to down sample the possible escape points, with candidate trajectories generated. Dynamically infeasible candidate trajectories are discarded. (D) Lowest cost, dynamically feasible trajectory is selected and sent to the controller. }
  \label{fig:stopping_trajectory}
\end{figure*}

\begin{figure}[b!]
  \centering
  \includegraphics[width=0.6\linewidth]{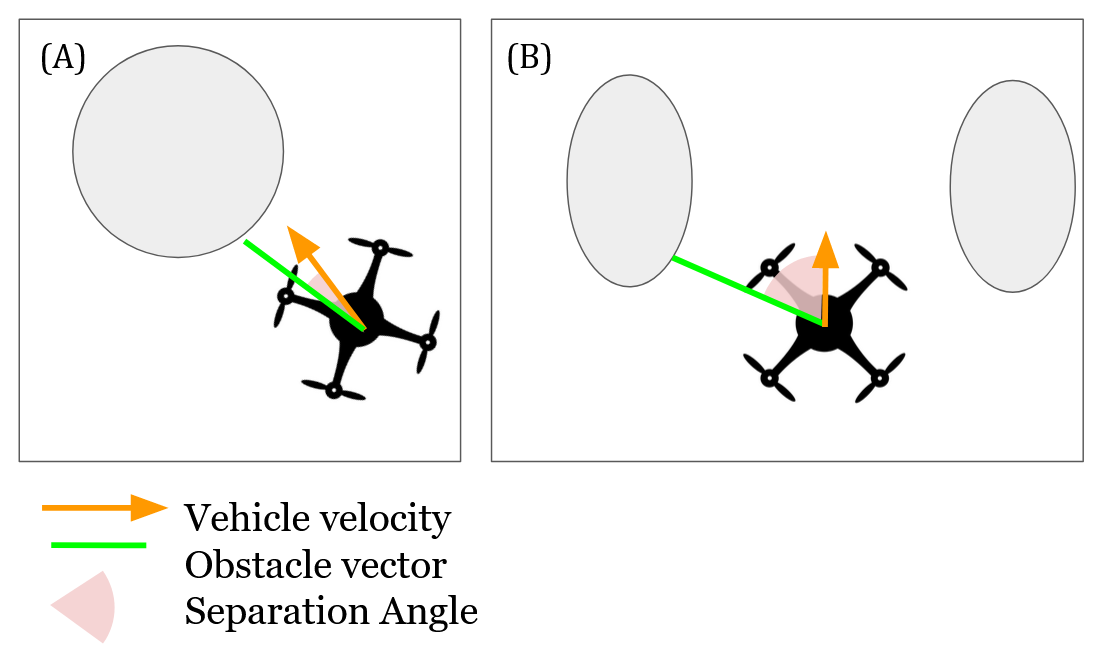}
  \caption{(A) Vehicle is heading in the direction of an obstacle. Vehicle is in higher risk of collision, $C_\text{stop}$ is lower. (B) Vehicle is heading between obstacles/ Vehicle is in lower risk of collision, $C_\text{stop}$ higher.} \label{fig:stop_angle}
\end{figure}

\subsection{Escape points sampling}
\input{escape_points.tex}

\subsection{Stopping trajectory generation}
\input{trajectory_generation.tex}

%% file: collision_monitoring.tex
We first find the set of possible collision points nearby by evaluating the normalized vector projection between a set of closest obstacle locations $\{\pv \}$ given by a known map and the vehicle velocity:
\begin{align*}
  \text{proj}(\xv, \pv) = \bigg \langle \frac{\mathbf{\dot{x}}}{\| \mathbf{\dot{x}} \|},
    \frac{\pv - \xv}{\| \pv - \xv \|} \bigg \rangle =
  \bigg \langle \frac{\mathbf{\dot{x}}}{\| \mathbf{\dot{x}} \|},
    \frac{\rv}{\| \rv \|} \bigg \rangle
\end{align*}
Where $\xv$, $\dot{\xv}$ is the vehicle position and velocity respectively, and the vector to an obstacle point $\pv$ is $\rv = \pv - \xv$.
The points away from the direction of motion, e.g., $\text{proj}(\xv, \pv){<} 0$ are discarded.

Then, for all points $\{ \pv \}$ in remaining set, we compute a combined stop criterion based on distance to the obstacle, speed, and the angle offset:
\begin{align*}
  C_\text{stop}(\xv, \pv) = w_1\norm{\rv} - w_2\norm{\dot{\xv}} + w_3\arccos({\text{proj}(\xv, \pv)})& \\
   \text{if} \; {\text{proj}(\xv, \pv)}& >= 0
\end{align*}

The third term represents the relative angle difference between the heading of the vehicle and the direction of the obstacle point, as shown in Figure \ref{fig:stop_angle}. 
A imminent stop trajectory is issued if for any obstacle point $\pv$, $C_\text{stop}(\xv, \pv) < \beta$.

%% file: fig_open.tex
\begin{figure}[t!]
  \centering
  \includegraphics[width=0.49\linewidth, trim=0 250px 0 0, clip]{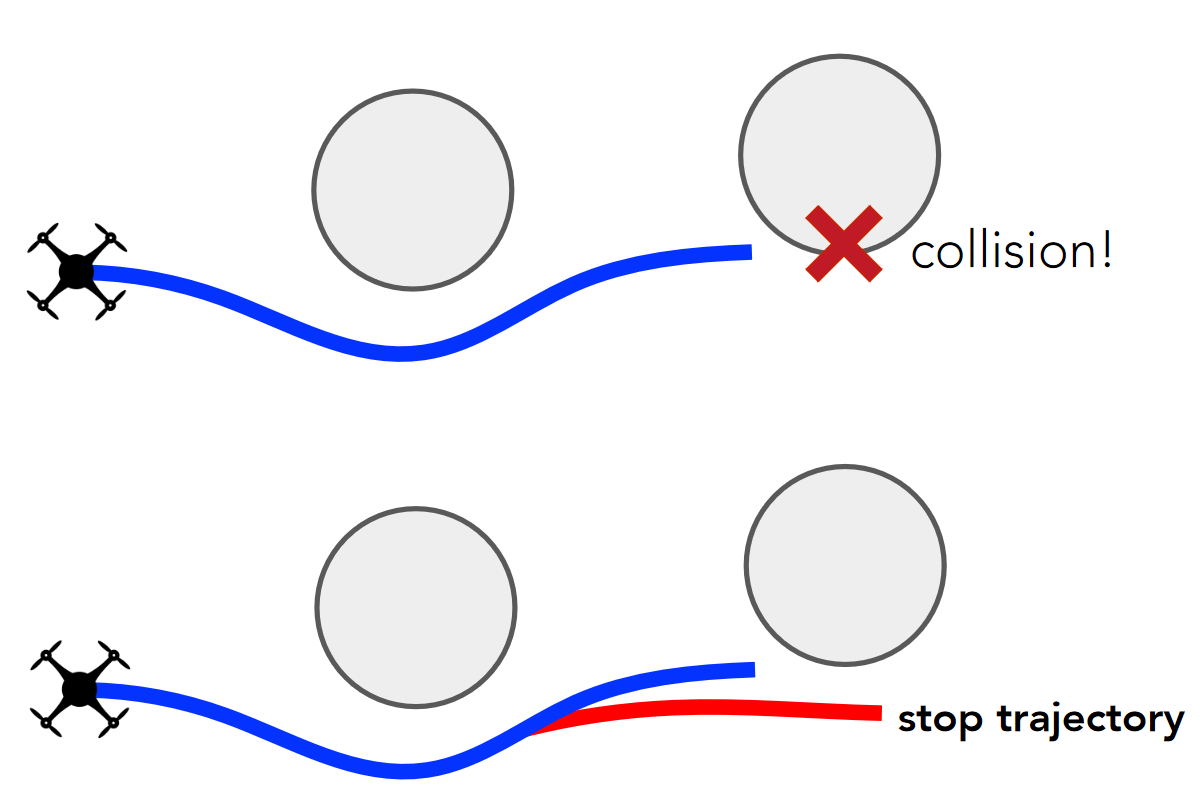}
  \includegraphics[width=0.49\linewidth, trim=0 0 0 270px, clip]{figures/clipart/stop_traj_problem_formulation.png}
  \caption{Trajectory-based teleoperation, current trajectory is collision free, but imminent collision after the trajectory.} \label{fig:formulation}
\end{figure}

%% file: escape_points.tex
Escape points are defined as possible goal points where the vehicle could stop. We first generate escape points and then plan trajectories to them. The first feasible trajectory incurring the lowest cost is sent to the controller. Thus, to minimize search time, we aim to check escape points that are more likely to produce feasible trajectories. 

An initial group of escape points $\{ \ev \}$ is generated using a uniform 3D grid, scaled proportionally to the vehicle's velocity. Points are sampled along the direction of motion using stratified sampling for points near the vehicle's original heading via the following cost:
$$C_\text{escape}(\xv, \ev) = w_1 q + w_2 d$$
Where $q{=}||(\xv {-} \ev {-}((\xv {-} \ev) {\cdot} \hat{\dot{\xv}})\hat{\dot{\xv}}||$ is the shortest distance from the escape point $\ev$ to the line $l(s) {=}\hat{\dot{\xv}}s{ +} \xv$ along the velocity vector, $\hat{\dot{\xv}}$ is the normalized velocity vector, and $d$ is the distance from the the escape point to its nearest obstacle.

First, points farther from obstacles are lower cost to encourage the vehicle towards an area with more free space. Second, points closer to the vehicle’s velocity vector are given lower costs.  Points farther from the vehicle’s path require the robot to turn severely, which results in undesirable dynamics.

Next, points are downsampled before trajectory generation via stratified sampling, which samples a fixed number of points from each strata based on cost. To ensure enough high quality points are chosen, more points are sampled from higher strata (low cost). In this implementation, 10 points are sampled from top 1\%, 40 points from next 9\%, 30 from next 40\% and 20 from bottom 50\%. The stratified method guarantees a set of points with both high and low costs such that in the event low cost trajectories are not dynamically feasible, we are able to sacrifice cost for dynamic feasibility.

%% file: trajectory_generation.tex
Given the set of escape points, we generate candidate stop trajectories starting with the lowest $C_\text{escape}$.

A trajectory is a time-parameterized function $\xi(t)$, defined over a time interval $t \in [0, T]$ that maps a given time $t$ to a position $\xv_t$.
A stop trajectory is a trajectory that brings a vehicle from in-motion to at-rest:
\begin{align*}
  \text{s.t.} \quad
  &\xi : [0, T] \rightarrow X \\
  &\xi^{(j)}(T) = 0, j > 0 \\
  &\xi \in \Xi_{\text{safe}}
\end{align*}

A single polynomial trajectory is computed by solving an 8th order polynomial on the current vehicle state and desired vehicle state. The current vehicle state is defined by the current position, velocity, acceleration, and jerk in each dimension ($x, y, z, yaw$). The desired state is the position of the escape point, with all higher-order derivatives of 0. Each dimension's polynomial is computed independently.

We solve for the coefficients, $c_i$, of the polynomial
$$ f(x) =\sum_{i=0}^{8}c_ix_i$$
with constraints
\begin{align*}
f(0) &= x_0, \dot{f}(0) = v_0, \ddot{f}(0) = a_0, f^{(3)}(0) = j_0\\
f(t) &=x_f, \dot{f}(t), \ddot{f}(t), f^{(3)}(t) =0
\end{align*}
where $x_0,v_0,a_0,j_0$ are the initial state and $x_f$ is the final position given by the escape point.

Next, the trajectory is interpolated and waypoints are checked against a known map to ensure it is collision-free. The trajectory is also checked for dynamic feasibility via an acceleration bound along the trajectory, i.e., $ \ddot\xv{<} \bar\alpha$. For our experiments, we select $\bar\alpha = 10\text{m/s}^2$.

The first trajectory that is safe and feasible is executed, and the search stops. If all potential escape points have been exhausted without finding a valid trajectory, then a stopping trajectory with no goal position is executed as a last effort.

\begin{algorithm}[b!]
  \caption{Stopping Trajectory Generation}\label{alg:traj_generation}
    \KwIn{Vehicle position $\xv$ and sample set of points $\mathcal{S}$}
    \For{$s \in \mathcal{S}$}{
        Compute $C_{escape}$\;
    }
    $S_{escape}$:=StratifiedSample($\mathcal{S}$)\;
    sort(escapePoints, $C_{escape})$\;

    \For{$\ev_i$ in $S_{escape}$}{
        $\xi$:=SolvePolynomial($\xv$, $\ev_i$)\;
        \If{feasible($\xi$) and collisionFree($\xi$)}
        {executeTrajectory($\xi$)\;}
    }
\end{algorithm}

%% file: fig_warehouse_run.tex
\begin{figure*}[t!]
    \centering
    \includegraphics[width=0.28\linewidth]{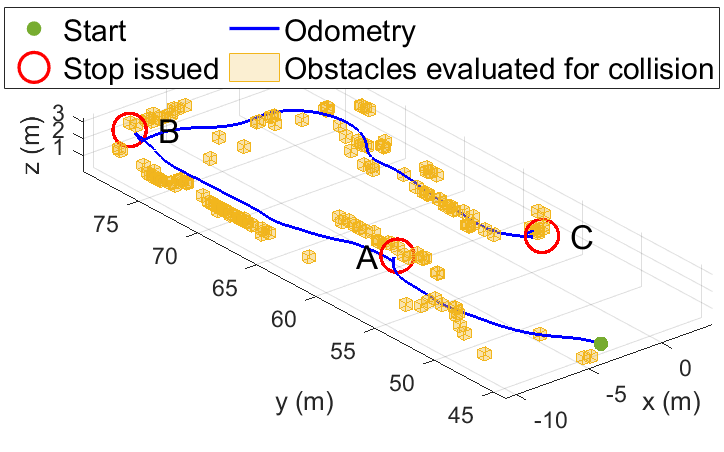}
    \includegraphics[width=0.7\linewidth]{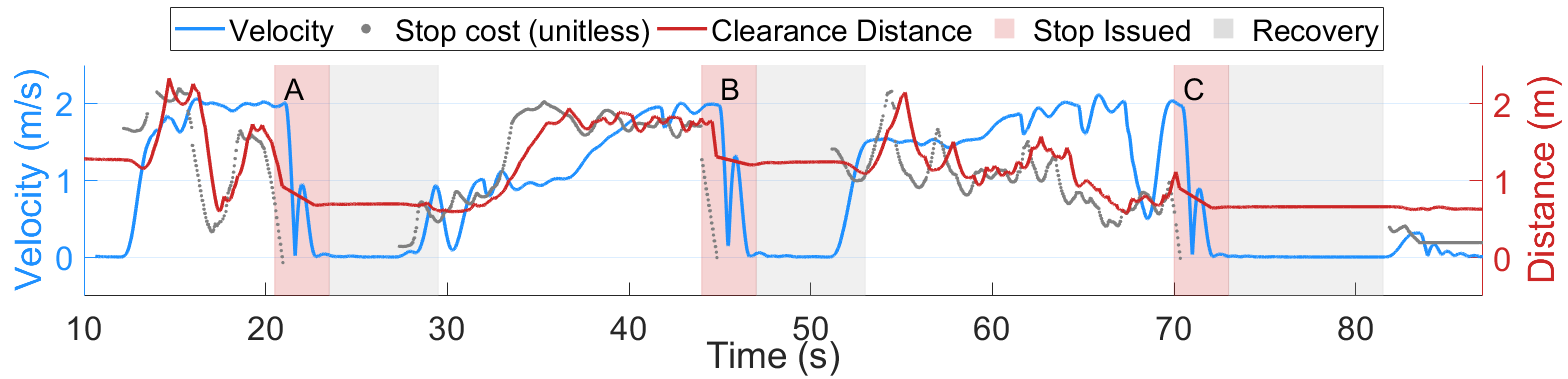}
    \caption{Trial run in the warehouse environment with imminent collision checking enabled. Left: Vehicle odometry with queried obstacles. Locations A, B, C indicate where stopping trajectories were issued corresponding with the left plot. Right: Stopping trajectories are issued when stop cost drops below a threshold, ensuring that the vehicle is safe. The operator recovers the vehicle with an in-place yaw before normal flight is resumed.}
    \label{fig:stopping_trial}
  \end{figure*}

%% file: experiments.tex
\section{Experiment and Results}

We evaluate our method in a simulated random forest environment and in a dense realistic warehouse environment in a teleoperated navigation task minimizing time. We conduct 20 trials in each environment, 10 with the proposed pipeline enabled, and 10 without. In our experiments, the stop criterion parameters $w_1$, $w_2$, $w_3$, $\beta$ are chosen to be $0.6$, $0.4$, $1.2$, $0.3$ respectively.

\begin{figure}[h!]
    \centering
    \includegraphics[width=\linewidth]{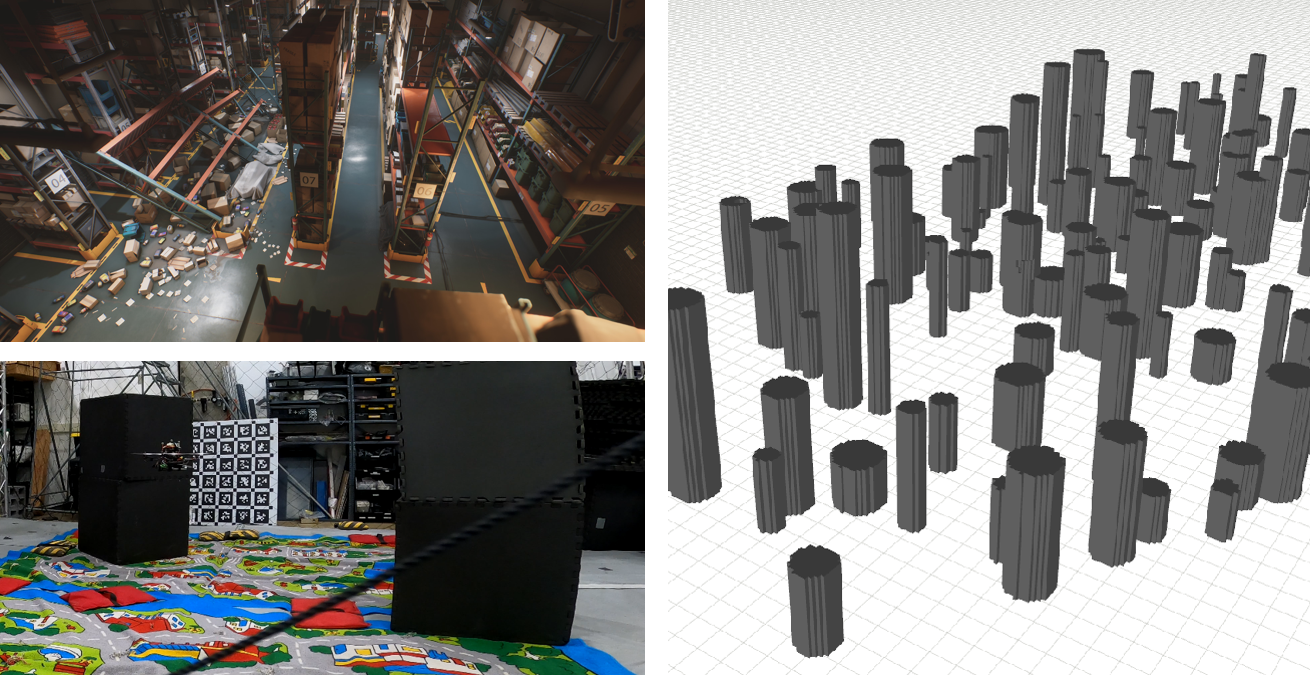}
    \caption{Testing scenarios. (top, left) Simulated warehouse environment. (bottom, left) Real life flight arena. (right) Simulated random forest environment.} \label{fig:env}
\end{figure}

We discuss task completion with and without enabling collision monitoring in Table \ref{tab:comparison}. With collision avoidance enabled, the operator completes the task with a success rate of 90\% for the forest scenario and 100\% for the warehouse scenario. When collision monitoring is disabled, the vehicle collides under aggressive flight, leading to 20\% task completion rate. The average velocity of enabled collision monitoring is comparable to disabled collision monitoring, but the vehicle safety is maintained. ``Distance'' refers to the average distance from obstacles throughout the entire trial, while ``Min Distance'' is the minimum distance from an obstacle on a given trial.
\input{tables/table_comparison.tex}

Table \ref{tab:num_stops} shows the average number of stops issued over the course of completing task with collision monitoring enabled. We prevent 6 potential collisions in the forest scenario on average, 4 collisions in the warehouse scenario and 9 collisions in the hardware experiment. The table shows the vehicle velocity and obstacle distance conditions that trigger these stops.
\input{tables/table_num_stops.tex}

In the hardware experiments, we teleoperate the vehicle in a flight arena with two pillars with imminent collision monitoring enabled. The vehicle flies between and around the pillars, but successfully stops if it approaches either pillar. 

\begin{figure}[h!]
    \centering
    \includegraphics[width=0.7\linewidth]{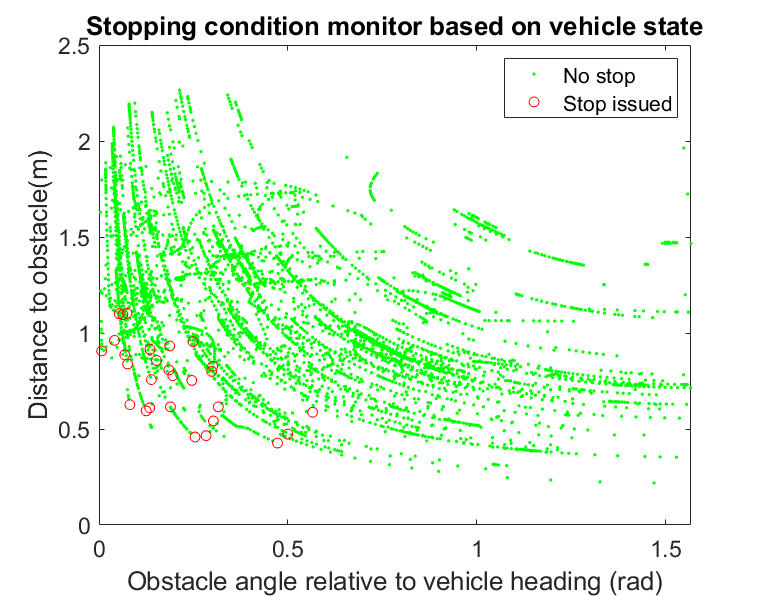}
    \caption{The stop criterion parameters include distance and relative angle to obstacle. An approximately linear boundary is formed between stop and no-stop conditions. Red circles represent conditions for which stops were triggered.}
    \label{fig:stopping_conditions}
\end{figure}

Figure \ref{fig:stopping_conditions} shows the correlation between the obstacle offset angle and the distance to the obstacle when determining whether to stop from hardware trials. A rougly linear boundary is formed between stop and no-stop scenarios. Some no-stop points lie below the boundary due to the third velocity factor, because the vehicle does not stop when it near obstacles at low velocity.

%% file: tables/table_comparison.tex
{
\setlength{\tabcolsep}{2pt} 
\begin{table}[t!]
  \centering
  \caption{Task completion with and without collision monitoring }  \label{tab:comparison}
  \begin{tabular} {l || c | c | c | c }
    \hline
    Scenario & Success rate & Velocity (m/s) & Distance (m) & Min Distance (m) \\
    \hline
    \textbf{Forest} & \\
    Enabled &  $ 90\% $   &  $ 1.37 \pm 0.18 $   &  $ 1.61 \pm 0.11 $ & $ 0.54 \pm 0.13$  \\
    Disabled &  $ 20\% $   &  $ 1.69 \pm 0.18 $   &  $ 1.44 \pm 0.22 $ & $ 0.39 \pm 0.16$  \\
    \textbf{Warehouse} & \\
    Enabled &  $ 100\% $   &  $ 1.73 \pm 0.07 $   &  $ 1.89 \pm 0.11 $ & $ 0.47 \pm 0.08$  \\
    Disabled &  $ 20\% $   &  $ 1.80 \pm 0.15 $   &  $ 1.82 \pm 0.20 $ & $ 0.32 \pm 0.13$  \\
    \hline
  \end{tabular}
\end{table}
}

%% file: tables/table_num_stops.tex
{
\setlength{\tabcolsep}{2pt} 
\begin{table}[t!]
  \centering
  \caption{Results for trials with collision detecting enabled }  \label{tab:num_stops}
  \begin{tabular} {l || c | c | c }
    \hline
    Environment & \# Stops Issued & Velocity (m/s) & Obstacle Distance (m) \\
    \hline
    Forest &  $ 6 \pm 1 $   &  $ 1.84 \pm 0.12 $   &  $ 1.39 \pm 0.09 $  \\
    Warehouse & $ 4 \pm 1 $   &  $ 1.97 \pm 0.04 $  &   $ 1.37 \pm 0.08 $ \\
    Hardware & $ 9 \pm 2 $ & $ 0.38 \pm 0.04 $ & $ 0.73 \pm 0.17 $ \\
    \hline
  \end{tabular}
\end{table}
}

%% file: conclusion.tex
\section{Conclusion and Future Work}
In this work, we presented an imminent collision avoidance system for safe teleoperation in dense environments. This system revolves around three main ideas. First, a continuous monitor on the vehicle state and nearby obstacles, which will trigger a stop if conditions are met. Second, escape point sampling to select potential goal locations in free space for the vehicle to stop.  Third, a polynomial trajectoy generator which iterates over escape points to select a feasibile and collision-free stopping trajectory. The system can be integrated in an existing teleoperation framework.

Future work can improve upon the stopping condition heuristic. This work used a KD-tree map representation, but could be expanded to use other map representations.

%% file: fig_forest_hw_run.tex
\begin{figure}[b!]
      \centering
      \includegraphics[width=0.45\linewidth]{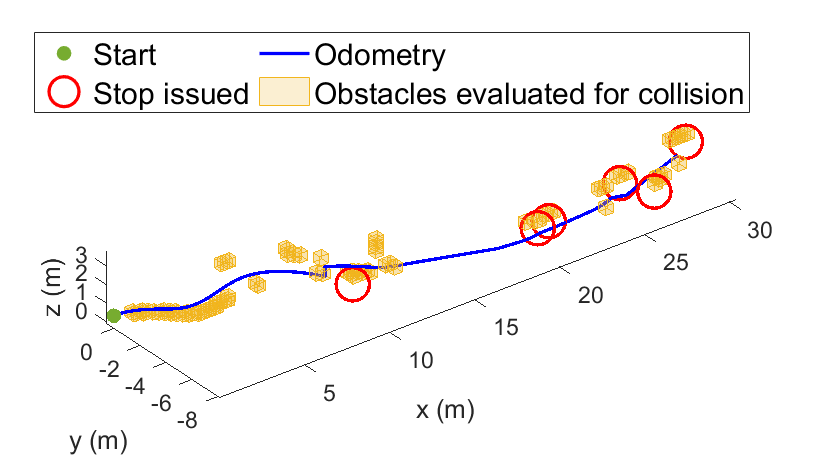}
      \includegraphics[width=0.45\linewidth]{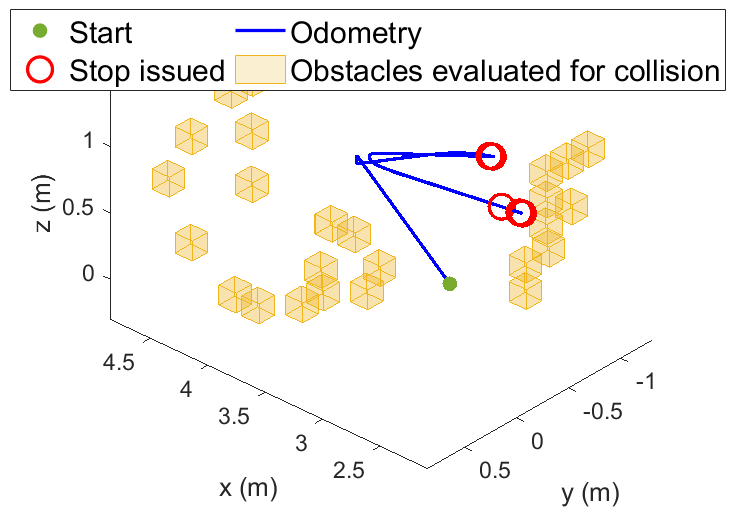}
    \caption{Trial run with imminent collision checking enabled in simulated random forest environment (Left) and real-life flight arena (Right). Vehicle odometry with queried obstacles, locations in red indicate where stopped trajectories were issued.}
    \label{fig:forest_run}
\end{figure}